\documentclass[10pt,twocolumn,letterpaper]{article}

\usepackage[pagenumbers]{cvpr} 
\usepackage{booktabs, multirow, multicol, array, caption, amsmath, amssymb, algorithm, algpseudocode }
\usepackage{graphicx}
\newcommand{\mz}[1]{\textcolor{black}{#1}}

\graphicspath{{./sec/images/}}

\definecolor{cvprblue}{rgb}{0.21,0.49,0.74}
\usepackage[pagebackref,breaklinks,colorlinks,allcolors=cvprblue]{hyperref}

\def\paperID{1012} 
\def\confName{CVPR}
\def\confYear{2026}

\title{FOUND: Fourier-based von Mises Distribution for Robust Single Domain Generalization in Object Detection}

\author{
Mengzhu Wang$^{1}$ \and 
Changyuan Deng$^{1}$ \and 
Shanshan Wang$^{2}$ \and 
Nan Yin$^{3}$ \and 
Long Lan$^{4}$ \and 
Liang Yang$^{1}\thanks{Corresponding author}$\\
{\small $^{1}$Hebei University of Technology \quad
 $^{2}$Anhui University \quad
 $^{3}$Hong Kong University of Technology \quad
 $^{4}$National University of Technology}\\
{\small \texttt{dreamkily@gmail.com}}
}

\begin{document}

\makeatletter
\g@addto@macro\@maketitle{
  \begin{figure}[H]
  \vspace{-0.5cm}
  \setlength{\linewidth}{\textwidth}
  \setlength{\hsize}{\textwidth}
  \centering
  \includegraphics[width=0.9\linewidth]{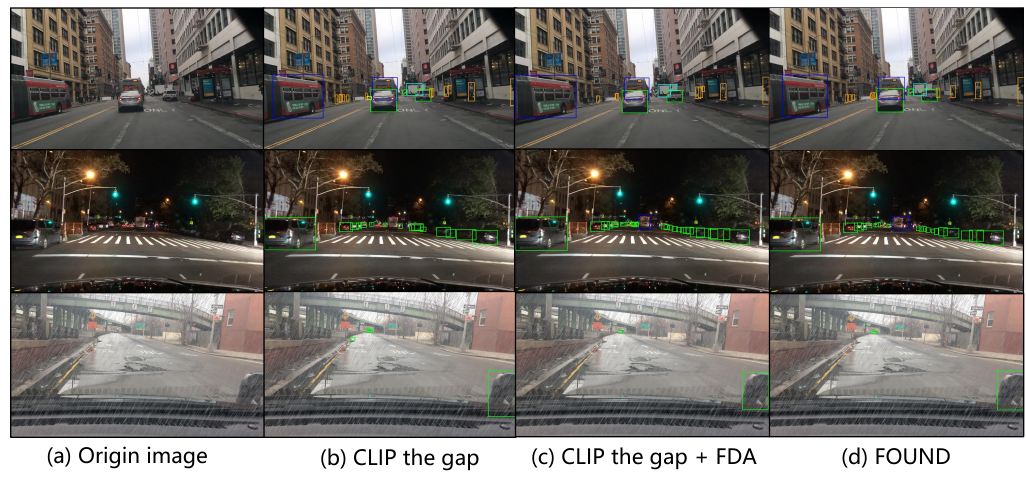}
  \caption{Overall of the framework. (a) As seen in the first row of images, CLIP the Gap makes two classification errors, misidentifying a mailbox on the left as a person and a pedestrian as a bicycle. In contrast, our method correctly identifies both targets. In the second row, CLIP the Gap fails to detect the bus, while our method accurately recognizes it. Furthermore, in the third row, CLIP the Gap generates a false positive by detecting a car in a background region, an error our method avoids.} 
  \label{fig:intro}
  \vspace{-0.0cm}
  \end{figure}
}
\makeatother

\maketitle
\begin{abstract}
\mz{Single Domain Generalization (SDG) for object detection aims to train a model on a single source domain that can generalize effectively to unseen target domains. While recent methods like CLIP-based semantic augmentation have shown promise, they often overlook the underlying structure of feature distributions and frequency-domain characteristics that are critical for robustness. In this paper, we propose a novel framework that enhances SDG object detection by integrating the von Mises-Fisher (vMF) distribution and Fourier transformation into a CLIP-guided pipeline. Specifically, we model the directional features of object representations using vMF to better capture domain-invariant semantic structures in the embedding space. Additionally, we introduce a Fourier-based augmentation strategy that perturbs amplitude and phase components to simulate domain shifts in the frequency domain, further improving feature robustness. Our method not only preserves the semantic alignment benefits of CLIP but also enriches feature diversity and structural consistency across domains. Extensive experiments on the diverse weather-driving benchmark demonstrate that our approach outperforms the existing state-of-the-art method.}
\end{abstract}

\section{Introduction}
\label{sec:intro}
\mz{Object detection~\cite{wang2024yolov10,zhang2025style,wu2025percept} has achieved significant success in controlled environments, yet its performance often degrades dramatically when faced with unseen domain shifts, such as changes in weather, lighting and scene style. This problem is typically addressed by Domain Adaptation (DA)~\cite{you2019universal,ben2006analysis,ganin2015unsupervised}, which adapts a model to a specific target domain. However, in many real-world applications, data from target domains is unavailable during training which may cause these methods inapplicable. This limitation has spurred interest in Domain Generalization (DG)~\cite{li2025seeking,agarwal2025tide}, which aims to learn models that can generalize to any unseen domain. A particularly challenging subset of this problem is Single Domain Generalization (SDG)~\cite{wang2021learning,qiao2020learning}, where the model is trained on data from only a single source domain.}

\mz{While SDG has been explored for image classification, its application to object detection remains relatively unexplored. Object detection introduces the compounded challenge of learning both domain-invariant representations for recognition and robust features for localization. A notable pioneering work Single-DGOD~\cite{Vidit_2023_CVPR} used disentanglement and self-distillation to learn domain-invariant features.  More recently, a novel approach PGST~\cite{li2025phrase} leveraged the pre-trained vision-language model CLIP to address this problem. However, CLIP typically provides visual representations at the image level and does not capture fine-grained visual representations at the object level for detection tasks. This limitation restricts the model's ability to perceive domain-invariant characteristics of individual instances  when dealing with significant appearance variations across domains. Furthermore, CLIP's semantic augmentation operates primarily in its joint embedding space, failing to fully exploit the structural patterns in the original pixel space that could effectively simulate domain shifts.}

\mz{To address these challenges, we propose \underline{F}\underline{O}\underline{U}rier-based von Mises Distributio\underline{N} for Robust \underline{D}omain Generalization in Object Detection (FOUND) through two complementary mechanisms: Fourier-based amplitude perturbation and von Mises-Fisher distribution-based feature regularization. FOUND first introduce Fourier-based Domain Augmentation (FDA), which operates directly in the frequency domain to simulate domain shifts. We decompose images into amplitude and phase spectra, recognizing that amplitude components primarily encode style characteristics while phase information preserves semantic structures. Through strategic amplitude mixing and perturbation, our method generates diverse training samples that emulate realistic domain variations such as weather changes and illumination differences. This frequency-space augmentation provides crucial pixel-level diversity that is difficult to achieve through existing feature-space methods. Meanwhile,  we develop von Mises-Fisher regularization (vMF) to enhance object-level representation learning.  Operating within CLIP's hyperspherical embedding space, we model object features using the vMF distribution to explicitly capture their directional properties. The concentration parameter $\kappa$ in vMF distribution enables precise control over feature dispersion on the unit hypersphere, facilitating the learning of compact and domain-invariant object representations. This geometric regularization ensures consistent directional relationships for features of the same category across domains, while improving angular separability between different categories.}

\mz{We evaluate our proposed FOUND framework on the challenging SDG driving benchmark introduced by \cite{wu2022single}, which features substantial domain shifts induced by diverse weather conditions and illumination changes. Comprehensive experiments demonstrate that our method successfully harmonizes Fourier-based augmentation with von Mises-Fisher regularization, not only compensating for the limitations of frequency-domain perturbation but fully unleashing its potential for domain generalization. Consequently, FOUND establishes new state-of-the-art performance across all unseen target domains. Our main contributions are as follows:}
\begin{itemize}
\item \mz{We introduce a novel Fourier-based Domain Augmentation (FDA) module that systematically perturbs amplitude spectra in the frequency domain to simulate realistic domain shifts, effectively expanding input-level diversity while preserving structural content in the phase component.}
\item \mz{We propose a von Mises-Fisher Regularization (vMF) framework that explicitly models object-level features on the hypersphere, enhancing domain invariance through directional concentration control and angular feature separation.}
\item \mz{We integrate these two complementary approaches into FOUND, a unified framework that achieves state-of-the-art performance on challenging SDG object detection benchmarks while demonstrating significant improvements in cross-domain generalization.}
\end{itemize}

\begin{figure*}[ht]
\centering
    \includegraphics[width=0.8\textwidth]{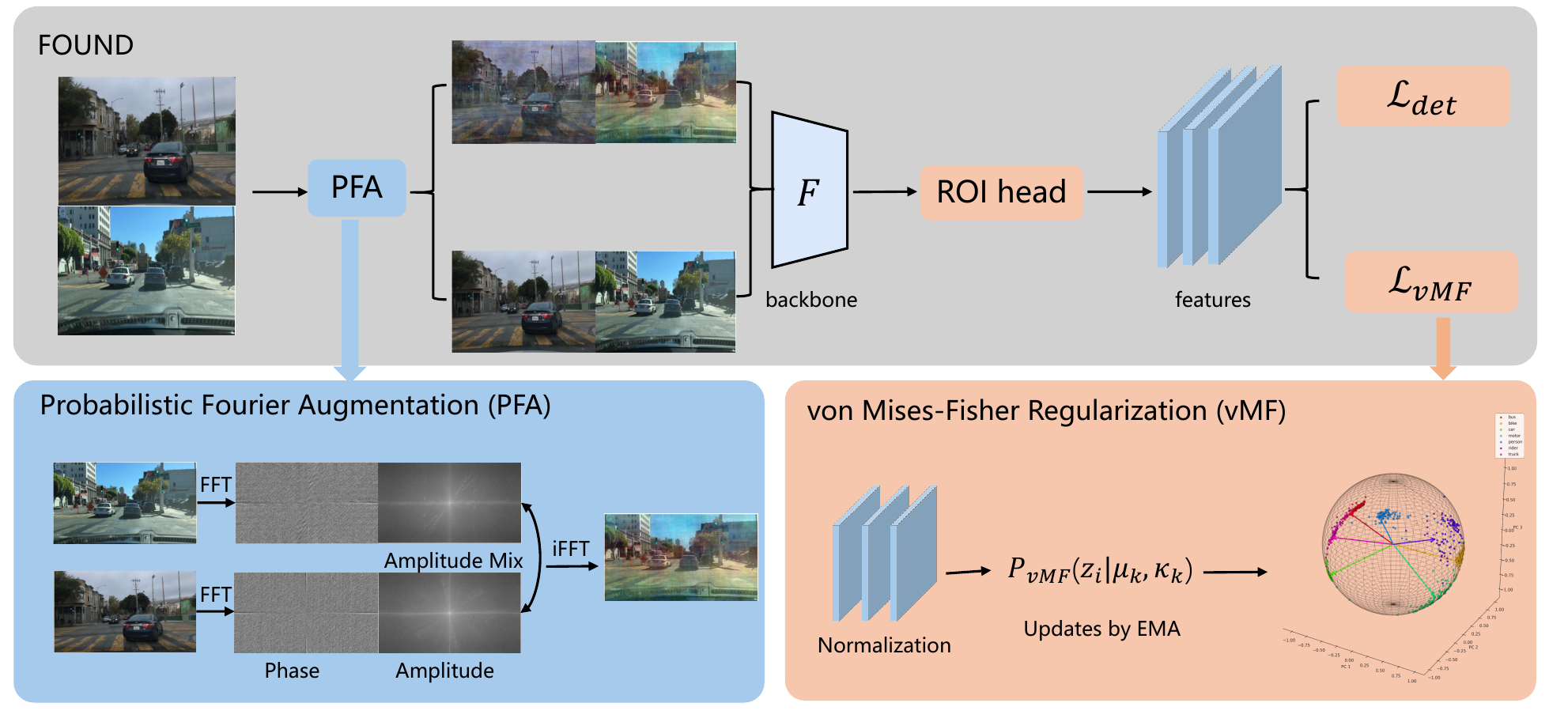}
    \caption{Overall framework the proposed FOUND,  the PFA module facilitates the learning of domain-invariant features, while the vMF module maintains semantic consistency and preserves diverse visual styles.}
    \label{fig:framework}
\end{figure*}

\section{Related Work}
\label{sec:related_work}

{\bf Single Domain Generalization (SDG).} Single Domain Generalization aims to learn models from a single source domain that are robust to unseen distributional shifts, a problem of critical importance as models often fail when deployed in novel environments~\cite{volpi2018generalizing,qiao2020learning,yuan2025asgs}. A dominant research direction is data augmentation. This ranges from stochastic transformations such as CutMix\cite{yun2019cutmix} and AugMix~\cite{hendrycks2019augmix} to adversarial methods that generate challenging out-of-distribution samples~\cite{volpi2018generalizing}.To this end, recent augmentation strategies, such as the work of Danish et al.\cite{danish2024improving}, focus on improving generalization by explicitly diversifying source data and then aligning the resulting feature distributions. Another significant approach involves architectural modifications, particularly through normalization layers. Methods like IBN-Net\cite{pan2018two} and ISW~\cite{choi2021robustnet} aim to disentangle domain-specific style from domain-invariant content. Following this principle, more explicit feature disentanglement frameworks have been developed, such as S-DGOD~\cite{wu2022single}, which uses a cyclic self-distillation process to isolate domain-invariant representations for the detector.More recently, Vision-Language Models (VLMs) like CLIP~\cite{pmlr-v139-radford21a} have been leveraged for their rich semantic priors. The seminal CLIP the Gap~\cite{Vidit_2023_CVPR} demonstrated using textual prompts for feature-space augmentation. However, these VLM-based approaches operate primarily in a high-level semantic space and often lack an explicit mechanism to enforce the structural consistency of learned features, a gap our work aims to fill by exploring frequency-domain properties and feature space geometry.

\noindent {\bf (Fast) Fourier Transform.} The Fourier transform's ability to decouple an image's phase (containing structural content) from its amplitude (containing stylistic statistics) is a well-established principle~\cite{Oppenheim1980TheIO}. This makes it a compelling tool for domain generalization. The core strategy involves creating augmented data by manipulating the amplitude spectrum while preserving the phase. A key example is the Fourier-based Framework for Domain Generalization~\cite{Xu_2021_CVPR}, which introduces Amplitude Mix to force the model to prioritize domain-invariant phase information. Similar ideas have proven effective in related areas such as style transfer~\cite{Huang2017ArbitraryST} and domain adaptation~\cite{Yang2020FDAFD}. However, we observe that these deterministic transformations can be overly aggressive. This crucial observation motivates our exploration of a probabilistic Fourier augmentation, designed to harness its diversification power while mitigating the potential for semantic degradation.

\noindent {\bf von Mises-Fisher(vMF) Distribution.} The von Mises-Fisher (vMF) distribution is a probability distribution on a hypersphere, making it a natural fit for modeling L2-normalized deep features. Its primary application in deep learning is feature space regularization, where it helps to enforce a more structured and semantically coherent representation. This goal is philosophically aligned with the objectives of Supervised Contrastive Learning~\cite{Khosla2020SupervisedCL}, which also seeks to improve the structure of feature space. A recent direct application is seen in the work of \cite{zhang2024fine}, who use the vMF distribution for regularization of semantic consistency. They modeled the features of each class as a distinct vMF distribution and used a KL-divergence loss to encourage intra-class compactness. While this is an effective technique for organizing a feature space, we find that its benefits are limited if the features on which it operates have been semantically degraded by a preceding augmentation step. This highlights the central challenge our work addresses: creating a synergistic framework where vMF regularization acts as a crucial stabilizer, enabling the model to safely learn from the powerful yet potentially disruptive Fourier-based augmentations.
\section{Method}
\label{sec:method}


\mz{This work addresses the challenge of SDG for object detection, which involves training a model on single source domain to perform robustly on multiple unseen target domain. To this end, we propose a \underline{F}\underline{O}\underline{U}rier-based von Mises Distributio\underline{N} for Robust Single \underline{D}omain Generalization in Object Detection (FOUND). FOUND consists of three main modules: VLM-guided target semantic, probabilistic fourier augmentation and von Mises-Fisher regularization. VML-guided target semantics leverages a Vision-Language Model (CLIP) to align semantic features between the source domain and potential target domains using textual prompts. Probabilistic Fourier augmentation generates diverse synthetic samples by modifying the frequency components of the source domain, simulating potential target shifts. von Mises-Fisher regularization ensures robust, domain-invariant feature representations by aligning the model's embeddings on the unit hypersphere. The overall framework of this work is shown in Figure.~\ref{fig:framework}.}

\subsection{VLM-Guided Target Semantics}
\mz{To guide the model's generalization, we first leverage the powerful semantic capabilities of a VLM~\cite{wang2025dynamic,kim2024vlm,wang2024q} such as CLIP~\cite{pmlr-v139-radford21a}. Building on the approach of CLIP the Gap~\cite{Vidit_2023_CVPR}, we exploit the joint image-text embedding space of CLIP to quantify the semantic differences between the source domain and potential target domains, which is crucial for domain generalization.}

\mz{Specifically, we define concise and domain-specific textual prompts that describe both source and potential target domains. For example, the source domain may be described as "a photo in clear daytime weather" ($p_s$), while a target domain might be described as "a photo in rainy dusk conditions" ($p_t$). These prompts capture the unique environmental and contextual differences between domains, which are important for learning domain-invariant features. We then use the pre-trained and frozen text encoder $T$ from CLIP to extract the embedding vectors of these textual descriptions, enabling the model to project both visual and textual information into a common semantic space.}
\begin{equation} q_s = T(p_s), \quad q_t = T(p_t) \end{equation}
where $q_s, q_t \in \mathbb{R}^d$ are high-dimensional representations in the shared semantic space. The difference between these vectors is defined as the \mz{semantic shift vector}, \begin{equation} \Delta q = q_t - q_s \end{equation} 
\mz{This vector $\Delta q$ encodes the core semantic shift required to transition from the source to the target domain (e.g., changes in lighting from "daytime" to "dusk" and weather from "clear" to "rainy"). It provides direct semantic guidance for our subsequent data augmentation and feature regularization, ensuring that the generalization process is directed towards meaningful target domains.}

\mz{While the semantic shift vector $\Delta q$ provides high-level guidance for bridging the source and unseen domains in the CLIP embedding space, domain generalization requires the model to handle diverse low-level distributional variations that often occur in real-world environments. These variations are difficult to capture purely through semantic embedding alignment. To enhance robustness against such visual domain shifts, we further introduce a Probabilistic Fourier Augmentation (PFA) module. Operating in the frequency domain, PFA stochastically perturbs the amplitude spectrum of source images to emulate diverse appearance changes while preserving the phase information that maintains semantic structure. This design complements the semantic-level guidance from $\Delta q$, enabling the model to generalize more effectively across unseen domains.}

\subsection{Probabilistic Fourier Augmentation (PFA)}
\label{sec:method-pfa}
\mz{Our motivation arises from a fundamental property of the Fourier transformation~\cite{parchami2016recent,tao2008sampling} that it can effectively decouple the structural and stylistic components of an image signal. Inspired by FACT~\cite{Xu_2021_CVPR}, we hypothesize that the phase spectrum predominantly captures high-level semantic and structural information that remains relatively stable across domains, while the amplitude spectrum encodes low-level appearance statistics such as style, texture, and illumination that are inherently domain-specific.}

\mz{Based on this assumption, an ideal data augmentation strategy for domain generalization should perturb the amplitude spectrum to simulate diverse styles while preserving the phase spectrum to maintain semantic integrity. This forces the model to disregard domain-specific low-level statistics and instead focus on the more generalizable information contained within the phase. Our augmentation strategy combines a frequency-domain manipulation technique with a per-sample probabilistic application policy. This is achieved by first transforming the image into the frequency domain using the 2D Discrete Fourier Transform (DFT), which we compute efficiently via the Fast Fourier Transform (FFT) algorithm. For a single-channel image $x$, the DFT is defined as:}
\begin{equation}
\label{eq:dft_2d}
\mathcal{F}(x)(u, v) = \sum_{h=0}^{H-1} \sum_{w=0}^{W-1} x(h, w) e^{-j2\pi \left(\frac{h}{H}u + \frac{w}{W}v\right)}
\end{equation}

\mz{This transformation is extended to color images by applying it independently to each of the R, G, and B channels. At its core, our method then employs an Amplitude Mix (AM) strategy~\cite{Xu_2021_CVPR} on the resulting frequency components to generate stylistically diverse samples. This involves creating a new amplitude spectrum by linearly interpolating between that of the content image and a randomly selected style-source, then combining it with the original phase spectrum to reconstruct the augmented image. Crucially, this process is applied independently to each image in a training batch with a probability $p_{aug}$. This stochastic approach is the cornerstone of our method, ensuring a balanced mixture of original and augmented data that maintains training stability while preventing the model from overfitting to specific augmentation artifacts.}

\begin{algorithm}[h]
\caption{FOUND}
\label{alg:found}
\begin{algorithmic}[1]
\Require Mini-batch of source images and labels: $B = \{(I_i, Y_i)\}$
\Require Full source dataset for sampling: $D_s$
\Require Augmentation probability $p_{\text{aug}}$, vMF loss weight $\lambda_{\text{vMF}}$
\Require vMF class parameters: prototypes $\{\mu_k\}$, concentrations $\{\kappa_k\}$
\Ensure Total loss for backpropagation: $L_{\text{total}}$

\Statex
\State {\bf // 1. Probabilistic Fourier Augmentation}
\State Initialize augmented batch $B' \leftarrow \emptyset$.
\For{each image $I_i$ in the batch $B$}
    \State Sample a random number $r$ from a uniform distribution U(0,1).
    \If{$r < p_{\text{aug}}$}
        \State Randomly sample a style image $I_{\text{style}}$ from $D_s$.
        \State Decompose content image: $A_i, P_i \leftarrow \text{FFT}(I_i)$.
        \State Decompose style image:  $A_{\text{style}}\leftarrow \text{FFT}(I_{\text{style}})$.
        \State Sample interpolation weight: $\lambda \sim \text{Beta}(\alpha, \alpha)$.
        \State Create new amplitude: $\hat{A}_{\text{aug}} \leftarrow (1 - \lambda)A_i + \lambda A_{\text{style}}$.
        \State Reconstruct image: $I'_i \leftarrow \text{IFFT}(\hat{A}_{\text{aug}}, P_i)$.  
    \Else
        \State $I'_i \leftarrow I_i$. 
    \EndIf
    \State Add $(I'_i, Y_i)$ to $B'$.
\EndFor

\Statex
\State {\bf // 2. Forward Pass and Loss Calculation}
\State Pass augmented batch $B'$ through detector to get predictions and foreground RoI features $Z_{fg} = \{(z_j, y_j)\}$.
\State $L_{\text{det}} \leftarrow \text{Calculate Detection Loss}(\text{Predictions}, Y)$.
\State {\bf // Apply vMF regularization}
\State $\mathcal{L}_{\text{vMF}} \leftarrow -\frac{1}{|Z_{fg}|}\sum_{(z_j, y_j=k) \in Z_{fg}} \log p_{\text{vMF}}(z_j | \mu_k, \kappa_k)$.
\Comment{Pulls each feature $z_j$ towards its class centroid $\mu_k$}

\Statex
\State {\bf // 3. Combine Losses for Final Objective}
\State $\mathcal{L}_{\text{total}} \leftarrow L_{\text{det}} + \lambda_{\text{vMF}} \mathcal{L}_{\text{vMF}}$.

\State \Return $\mathcal{L}_{\text{total}}$
\end{algorithmic}
\end{algorithm}

\subsection{von Mises-Fisher Regularization}
\label{sec:method-vmf}
\mz{While PFA enhances data diversity, it may inadvertently destabilize the learned feature space. To mitigate this issue, we introduce a semantic consistency regularization scheme based on the von Mises-Fisher (vMF) distribution~\cite{banerjee2005clustering,du2024probabilistic,du2024simpro}. The vMF distribution provides a natural framework for modeling L2-normalized deep features on a hypersphere. Our objective is to maintain compact and semantically consistent feature representations for each class in the embedding space, irrespective of stylistic variations.}

\mz{Our theoretical foundation of this regularization lies in minimizing the KL divergence~\cite{van2014renyi,cui2024decoupled} between the empirical distribution of a feature vector modeled as a Dirac delta function and a target vMF distribution corresponding to its class. Following the insight of~\cite{zhang2024fine}, we leverage the well-established principle that minimizing this KL divergence is equivalent to maximizing the log-likelihood of the feature under the vMF distribution. Consequently we formulate our loss as the Negative Log-Likelihood (NLL) of the observed features:}
\begin{equation}
\mathcal{L}_{\text{vMF}} = -\mathbb{E}_{(z_i, y_i=k)}[\log p_{\text{vMF}}(z_i|\mu_k, \kappa_k)]
\end{equation}
\mz{where $p_{\text{vMF}}(z|\mu, \kappa) = C_d(\kappa) e^{\kappa \mu^T z}$ is the probability density function of the vMF distribution, and $C_d(\kappa)$ is a normalization constant dependent on the feature dimension and concentration. The distribution parameters for each class $k$ the mean direction (class centroid) $\mu_k$ and the concentration $\kappa_k$ are dynamically maintained via an Exponential Moving Average (EMA), allowing them to adapt smoothly throughout training.}


\mz{This loss term pulls all features of the same class toward their dynamically updated class centroid $\mu_k$ regardless of whether they come from original or augmented samples. The regularization works in synergy with PFA as PFA increases data diversity to promote generalization while the vMF loss preserves semantic structure and ensures the model maintains strong discriminative capability across diverse visual styles.}

\subsection{Overall Formulation}
\mz{Our final training objective combines the standard detection loss with the proposed feature regularization, optimized on data augmented by PFA. The total loss $\mathcal{L}_{\text{total}}$ is formulated as follow:}
\begin{equation}
\mathcal{L}_{\text{total}} = \mathcal{L}_{\text{det}} + \lambda_{\text{vMF}} \mathcal{L}_{\text{vMF}}
\label{eq:total_loss}
\end{equation}
\mz{where $\lambda_{\text{vMF}}$ is a balancing hyperparameter. The detection loss $\mathcal{L}{\text{det}}$ is adopted from our baseline CLIP Gap~\cite{Vidit_2023_CVPR} and handles the core tasks of object localization and classification. The proposed von Mises-Fisher regularization loss $\mathcal{L}{\text{vMF}}$ enforces semantic compactness in the learned RoI features and provides stability against the feature space dispersion caused by strong augmentations in PFA. The training process of FOUND is presented in Algorithm~\ref{alg:found}.}

\section{Experiments}
\label{sec:experiments}

\subsection{Experimental setup}
\mz{In this paper, our experiments are performed on a dataset that includes various challenging weather conditions. To ensure the fairness and comparability of our evaluation, we used the same dataset and division method as a previous study~\cite{Vidit_2023_CVPR}. The images in this dataset come from several well-known public datasets, including Berkeley Deep Drive 100K (BDD-100K)~\cite{yu2020bdd100k}, Cityscapes~\cite{cordts2016cityscapes} and its synthetic version Foggy City scapes~\cite{sakaridis2018semantic}, and the Adverse-Weather~\cite{hassaballah2020vehicle} dataset. Among them, some of the rainy and foggy images are generated from real images using rendering or synthesis techniques.}

\mz{Our model is trained only on the source domain, which is the daytime clear scene with 19,395 training images. Its generalization ability is then tested on four unseen target domains, including: 26,158 night clear images, 3,501 dusk rainy images, 2,494 night rainy images, and 3,775 daytime foggy images. This strict "train on clear, test on all-weather" setting can effectively test the model's robustness when facing big changes in light and weather. The dataset provides labels for seven common traffic object categories, such as cars, persons, and trucks.}

\subsection{Implementation details}
\mz{We utilize CLIP-the-Gap~\cite{Vidit_2023_CVPR} as our baseline model. Building upon this, we introduce two core components: 1) Probabilistic Fourier Augmentation (PFA), specifically implemented as Amplitude Mix; and 2) a von Mises-Fisher regularization loss (vMF) applied to the detector's RoI head. In addition to these modifications, all other training settings, such as the optimizer (SGD), the learning rate, and the batch size, are kept identical to the baseline model. For PFA, the frequency-domain augmentation is applied online during training, with each input image augmented with a 50\% probability using the Amplitude Mix strategy~\cite{Xu_2021_CVPR}. For a given content image, a style reference image is randomly sampled from the source training dataset. We perform a 2D Fast Fourier Transform (FFT) on both images to separate their amplitude and phase spectra. A new amplitude spectrum is formed by linearly interpolating between the amplitudes of the content and style images, controlled by a coefficient $\alpha$drawn from a uniform distribution $\mathcal{U}(0, 1.0)$. This new amplitude is then combined with the original phase spectrum of the content image, and the final augmented image is reconstructed via an inverse FFT. For vMF, to enforce semantic compactness in the feature space, we introduce a vMF regularization loss, which is applied to the 512-dimensional features extracted from foreground proposals in the RoI head. This loss is incorporated into the total training objective with a fixed weight of $0.005$. The distribution parameters are updated dynamically. The prototype direction vector $\mu$ for each class is maintained using an Exponential Moving Average (EMA) with a momentum of $0.99$. The concentration parameter $\kappa$ for each class is treated as a standard learnable parameter and is optimized via backpropagation.}

\subsection{Comparison with the State-of-the-Art}
\begin{table}[t]
\centering
\small
\setlength{\tabcolsep}{5pt} 
\begin{tabular}{@{}lccccc@{}}
\toprule
\textbf{Method} & 
\begin{tabular}{@{}c@{}}\textbf{Day} \\ \textbf{Clear}\end{tabular} & 
\begin{tabular}{@{}c@{}}\textbf{Night} \\ \textbf{Clear}\end{tabular} & 
\begin{tabular}{@{}c@{}}\textbf{Dusk} \\ \textbf{Rainy}\end{tabular} & 
\begin{tabular}{@{}c@{}}\textbf{Night} \\ \textbf{Rainy}\end{tabular} & 
\begin{tabular}{@{}c@{}}\textbf{Day} \\ \textbf{Foggy}\end{tabular} \\
\midrule
FR~\cite{ren2015faster} & 48.1 & 34.4 & 26.0 & 12.4 & 32.0 \\
SW~\cite{pan2019switchable} & 50.6 & 33.4 & 26.3 & 13.7 & 30.8 \\
IBN-Net~\cite{pan2018two} & 49.7 & 32.1 & 26.1 & 14.3 & 29.6 \\
IterNorm~\cite{huang2019iterative} & 43.9 & 29.6 & 22.8 & 12.6 & 28.4 \\
ISW~\cite{choi2021robustnet} & 51.3 & 33.2 & 25.9 & 14.1 & 31.8 \\
S-DGOD~\cite{wu2022single} & \textbf{56.1} & 36.6 & 28.2 & 16.6 & 33.5 \\
SRCD~\cite{rao2024srcd} & - & 36.7 & 28.8 & 17.0 & 35.9 \\
CLIP gap~\cite{Vidit_2023_CVPR} & 51.3 & \underline{36.9} & \underline{32.3} & \underline{18.7} & \underline{38.5} \\
\hline
\textbf{FOUND} & \underline{53.5} & \textbf{39.1} & \textbf{35.4} & \textbf{20.8} & \textbf{38.9} \\
\bottomrule
\end{tabular}
\caption{Performance comparison across different weather conditions. The best and suboptimal results are highlighted in bold font and with underlining, respectively.}
\label{table:res}
\end{table}
\mz{We evaluate our proposed framework, by comparing it against a diverse set of baselines and alternative methods to assess its effectiveness in single domain generalization. For this SDG task, our primary evaluation metric is the mean Average Precision (mAP) on four unseen target domains: daytime-foggy, night-rainy, dusk-rainy, and night-clear. We follow the standard protocol established by Single-DGOD and use the training-domain validation strategy for model selection.}

\mz{As shown in Table~\ref{table:res}, our method consistently outperforms the baseline across all domains. It demonstrates particularly strong performance under challenging night-time conditions, improving mAP by 2.1\% on night-rainy scenes and 2.2\% on night-clear scenes. A notable 3.1\% gain is also observed in the dusk-rainy domain. In contrast, on the daytime-foggy domain, our method maintains performance comparable to the baseline, effectively mitigating the degradation typically caused by aggressive augmentations. Moreover, the framework achieves higher accuracy on the source domain, confirming its ability to enhance cross-domain generalization without compromising in-domain learning.}

\begin{table}[h]
\centering
\footnotesize
\setlength{\tabcolsep}{3pt}
\begin{tabular}{@{}lcccccccc@{}}
\toprule
 & \multicolumn{7}{c}{AP} & mAP \\
\cmidrule(lr){2-8}
Method & Bus & Bike & Car & Motor & Person & Rider & Truck & All \\
\midrule
FR~\cite{ren2015faster} & 28.1 & 29.7 & 49.7 & 26.3 & 33.2 & 35.5 & 21.5 & 32.0 \\
S-DGOD~\cite{wu2022single} & 32.9 & 28.0 & 48.8 & 29.8 & 32.5 & 38.2 & 24.1 & 33.5 \\
SRCD~\cite{rao2024srcd} & 6.4 & 30.1 & 52.4 & 31.8 & 33.4 & 40.1 & 27.7 & 35.9 \\
CLIP gap~\cite{Vidit_2023_CVPR} & \underline{36.1} & \textbf{34.3} & \underline{58.0} & \underline{33.1} & \underline{39.0} & \textbf{43.9} & \underline{25.1} & \underline{38.5} \\
\midrule
FOUND & \textbf{37.4} & \underline{33.0} & \textbf{58.9} & \textbf{33.9} & \textbf{40.6} & \underline{43.2} & \textbf{25.8} & \textbf{38.9} \\
\bottomrule
\end{tabular}
\caption{Performance comparison across different object categories in the daytime clear to daytime foggy scene. The best and suboptimal results are highlighted in bold font and with underlining, respectively.}
\label{dayfoggy1}
\end{table}
\noindent {\bf Daytime Clear to Day Foggy.} \mz{As shown in Table~\ref{dayfoggy1}, object appearance changes drastically in foggy images compared to the day-clear scenario. FOUND achieves highly competitive performance comparable to the baseline, demonstrating its effectiveness in addressing the loss of contrast and high-frequency details caused by fog. This result is particularly noteworthy, as it confirms that our probabilistic approach successfully mitigates the catastrophic performance drop observed with deterministic Fourier augmentation, turning it into a beneficial regularization mechanism.}

\begin{table}[h]
\centering
\footnotesize
\setlength{\tabcolsep}{3pt}
\begin{tabular}{@{}lcccccccc@{}}
\toprule
 & \multicolumn{7}{c}{AP} & mAP \\
\cmidrule(lr){2-8}
Method & Bus & Bike & Car & Motor & Person & Rider & Truck & All \\
\midrule
FR~\cite{ren2015faster} & 34.7 & 32.0 & 56.6 & 13.6 & 37.4 & 27.6 & 38.6 & 34.4 \\
S-DGOD~\cite{wu2022single} & 40.6 & \underline{35.1} & 50.7 & 19.7 & 34.7 & \textbf{32.1} & \textbf{43.4} & 36.6 \\
SRCD~\cite{rao2024srcd} & 43.1 & 32.5 & 52.3 & \underline{20.1} & 34.8 & 31.5 & \underline{42.9} & 36.7 \\
CLIP gap~\cite{Vidit_2023_CVPR} & \underline{37.7} & 34.3 & \underline{58.0} & 19.2 & \underline{37.6} & 28.5 & \underline{42.9} & \underline{36.9} \\
\midrule
FOUND & \textbf{41.7} & \textbf{35.9} & \textbf{60.1} & \textbf{22.8} & \textbf{40.7} & \underline{30.7} & \textbf{43.4} & \textbf{39.3} \\
\bottomrule
\end{tabular}
\caption{Performance comparison across different object categories in the day clear to night clear scene. The best and suboptimal results are highlighted in bold font and with underlining, respectively.}
\label{tab:nightclear}
\end{table}
\noindent {\bf Daytime Clear to Night Clear.} \mz{As shown in Table~\ref{tab:nightclear}, the Night-Clear domain evaluates the model’s adaptability to illumination-only shifts. Our method significantly surpasses the baseline in this setting, validating the effectiveness of the frequency-domain augmentation. By directly perturbing the amplitude spectrum, PFA enables the model to achieve stronger invariance to lighting variations than feature-space semantic augmentation.}

\begin{table}[h]
\centering
\footnotesize
\setlength{\tabcolsep}{3pt}
\begin{tabular}{@{}lcccccccc@{}}
\toprule
 & \multicolumn{7}{c}{AP} & mAP \\
\cmidrule(lr){2-8}
Method & Bus & Bike & Car & Motor & Person & Rider & Truck & All \\
\midrule
FR~\cite{ren2015faster} & 28.5 & 20.3 & 58.2 & 6.5 & 23.4 & 11.3 & 33.9 & 26.0 \\
S-DGOD~\cite{wu2022single} & 37.1 & 19.6 & 50.9 & 13.4 & 19.7 & 16.3 & 40.7 & 28.2 \\
SRCD~\cite{rao2024srcd} & \underline{39.5} & 21.4 & 50.6 & 11.9 & 20.1 & 17.6 & 40.5 & 28.8 \\ 
CLIP gap~\cite{Vidit_2023_CVPR} & 37.8 & \underline{22.8} & \underline{60.7} & \underline{16.8} & \underline{26.8} & \underline{18.7} & \underline{42.4} & \underline{32.3} \\
\midrule
FOUND & \textbf{41.4} & \textbf{28.9} & \textbf{61.9} & \textbf{21.3} & \textbf{29.7} & \textbf{19.1} & \textbf{45.9} & \textbf{35.4} \\
\bottomrule
\end{tabular}
\caption{Performance comparison across different object categories in the day clear to dusk rainy scene. The best and suboptimal results are highlighted in bold font and with underlining, respectively.}
\label{table:duskrainy}
\end{table}
\noindent {\bf Daytime Clear to Dusk Rainy.} \mz{As shown in Table~\ref{table:duskrainy}, Dusk Rainy scenes present a compounded challenge characterized by both low illumination and complex rain-induced noise patterns. Our method achieves a noticeably larger performance improvement over the baseline in this scenario, highlighting the complementary strengths of its two core components. The proposed PFA effectively simulates the low contrast of dusk and the high-frequency perturbations introduced by rain, while the vMF regularization maintains compact and well-separated class representations.} 

\begin{table}[h]
\centering
\footnotesize
\setlength{\tabcolsep}{3pt}
\begin{tabular}{@{}lcccccccc@{}}
\toprule
 & \multicolumn{7}{c}{AP} & mAP \\
\cmidrule(lr){2-8}
Method & Bus & Bike & Car & Motor & Person & Rider & Truck & All \\
\midrule
FR~\cite{ren2015faster} & 16.8 & 6.9 & 26.3 & 0.6 & 11.6 & 9.4 & 15.4 & 12.4 \\
S-DGOD~\cite{wu2022single} & 24.4 & 11.6 & 29.5 & \textbf{9.8} & 10.5 & 11.4 & 19.2 & 16.6 \\
SRCD~\cite{rao2024srcd} & 26.5 & \underline{12.9} & 32.4 & 0.8 & 10.2 & \textbf{12.5} & \underline{24.0} & 17.0 \\
CLIP gap~\cite{Vidit_2023_CVPR} & \underline{28.6} & 12.1 & \underline{36.1} & 9.2 & \underline{12.3} & 9.6 & 22.9 & \underline{18.7} \\
\midrule
FOUND & \textbf{31.4} & \textbf{13.6} & \textbf{39.5} & \underline{9.4} & \textbf{14.4} & \underline{11.6} & \textbf{26.1} & \textbf{20.8} \\
\bottomrule
\end{tabular}
\caption{Performance comparison across different object categories in the day clear to night rainy scene. The best and suboptimal results are highlighted in bold font and with underlining, respectively.}
\label{tab:nightrainy}
\end{table}

\begin{table*}[t]
\centering
\footnotesize
\setlength{\tabcolsep}{8pt}
\begin{tabular}{@{}cccccccccc@{}}
\toprule
\multicolumn{3}{c}{Components} & \multicolumn{6}{c}{\textbf{mAP}} \\
\cmidrule(lr){1-3}\cmidrule(l){4-9}
\textbf{Original} & \textbf{PFA} & \textbf{vMF reg.} & 
\begin{tabular}{@{}c@{}}\textbf{Day} \\ \textbf{Clear}\end{tabular} & 
\begin{tabular}{@{}c@{}}\textbf{Night} \\ \textbf{Clear}\end{tabular} & 
\begin{tabular}{@{}c@{}}\textbf{Dusk} \\ \textbf{Rainy}\end{tabular} & 
\begin{tabular}{@{}c@{}}\textbf{Night} \\ \textbf{Rainy}\end{tabular} & 
\begin{tabular}{@{}c@{}}\textbf{Day} \\ \textbf{Foggy}\end{tabular} & 
\textbf{Avg.} \\
\midrule
\checkmark & — & — & 51.3 & 36.9 & 32.3 & 18.7 & 38.5 & 35.5 \\
\checkmark & $p=1.0$ & — & 51.7 & 35.9 & 34.4 & 20.0 & 33.2 & 35.0 \\
\checkmark & $p=1.0$ & \checkmark & 52.2 & 36.0 & 33.4 & 18.3 & 35.2 & 35.0 \\
\checkmark & $p=0.5$ & \checkmark & \textbf{53.5} & \textbf{39.1} & \textbf{35.4} & \textbf{20.8} & \textbf{38.9} & \textbf{37.5} \\
\bottomrule
\end{tabular}
\caption{Ablation study of different model components on cross-domain performance.}
\label{tab:ablation}
\end{table*}

\noindent {\bf Daytime Clear to Night Rainy.} \mz{As shown in Table~\ref{tab:nightrainy}, this represents the most challenging scenario, where dark night conditions are further complicated by rain-induced artifacts. Despite the extreme difficulty, our method achieves a clear improvement over the baseline, demonstrating its enhanced robustness. We hypothesize that under such near-adversarial conditions, the baseline model’s feature representations become unstable, leading to degraded performance. In contrast, the proposed vMF regularization serves as an effective stabilizer, preserving the structure of the feature space and ensuring consistent discrimination across severely shifted domains.}

\subsection{Ablation Study}
\mz{To analyze our proposed framework and validate the necessity of each design choice, we conduct a step-by-step ablation study. Our analysis is structured to mirror our research journey: from identifying the inherent pitfalls of a promising technique to developing a robust solution. We begin with the strong CLIP the Gap model as our baseline and incrementally introduce our components. (1) We first apply deterministic Fourier augmentation ($p=1.0$) to empirically demonstrate the phenomenon of semantic feature collapse and its detrimental impact on model performance. (2) We then incorporate the proposed vMF regularization to examine its role as a feature stabilizer, showing that although it alleviates the problem, it does not provide a complete solution. (3) Finally, we introduce a probabilistic augmentation strategy ($p=0.5$), illustrating that stochasticity is the key factor that transforms aggressive augmentation into an effective online regularizer, thereby fully realizing its potential for improving generalization.}


\mz{As shown in Table~\ref{tab:ablation}, we begin by applying full Fourier augmentation, performing amplitude mixing between two images from the single source domain to generate new samples while keeping all other settings identical to the baseline. Although this deterministic augmentation ($p=1.0$) yields slight improvements on certain target domains, it also causes catastrophic performance degradation on others. In particular, the mAP on the daytime-foggy domain drops sharply by 5.3 points (from 38.5\% to 33.2\%), with a similar decline observed on night-clear. This empirically confirms our core hypothesis that an unrestrained, deterministic application of strong augmentations can severely disrupt the model’s ability to learn consistent semantic representations which we refer to as semantic feature collapse.}

\mz{Having diagnosed the problem as a loss of feature space structure, we subsequently introduce von Mises-Fisher (vMF) regularization as a feature stabilizer. As shown in the third row of Table 6, incorporating the vMF loss effectively mitigates the damage, improving the daytime-foggy performance from 33.2\% to 35.2\% mAP. This partial recovery demonstrates that the vMF regularization counteracts the feature dispersion induced by aggressive augmentations by enforcing intra-class compactness in the embedding space.}

\mz{The final and most crucial improvement comes from introducing stochasticity into the augmentation process. By shifting from deterministic ($p=1.0$) to probabilistic ($p=0.5$) application of the vMF-regularized Fourier augmentation, our model achieves a remarkable boost in overall generalization. As shown in the last row of Table~\ref{tab:ablation},  the mAP on daytime-foggy fully recovers to 38.9\%, performing on par with the baseline and thus completely neutralizing the augmentation's negative side effects. More importantly, with the training dynamics now balanced, the generalization benefits of Fourier augmentation are fully unlocked across all other domains, leading to significant gains of +2.1\% mAP on night-rainy and +3.1\% mAP on dusk-rainy, respectively.}


\begin{figure}[tb]
\centering
   \includegraphics[width=\linewidth]{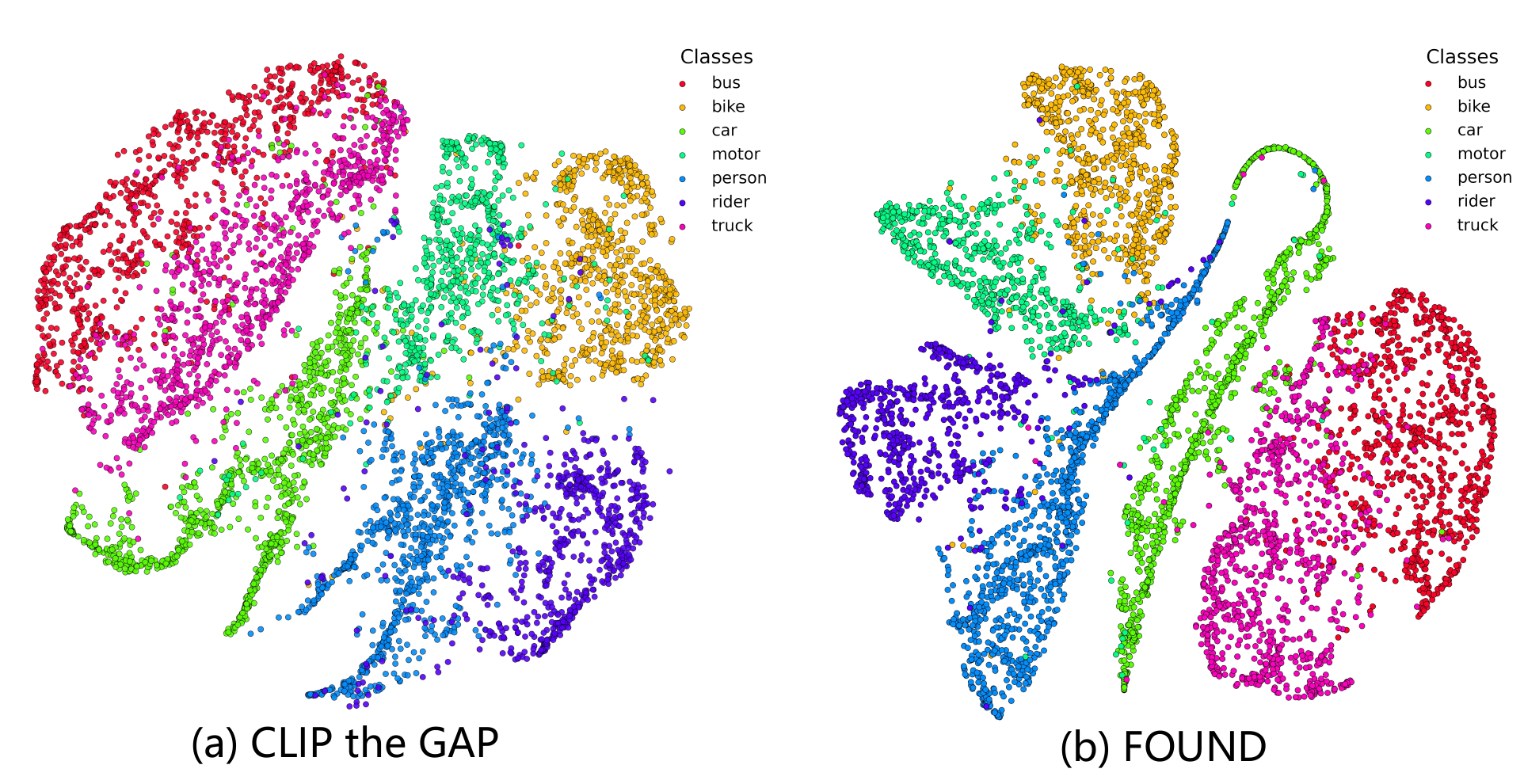}
   \caption{Visualization of the feature spaces for (a) CLIP the GAP and (b) FOUND. Compared to the entangled feature distribution of the baseline, FOUND learns a more separable and semantically aligned representation.}
   \label{fig:tsne}
\end{figure}

\begin{figure}[tb]
\centering
    \includegraphics[width=\linewidth]{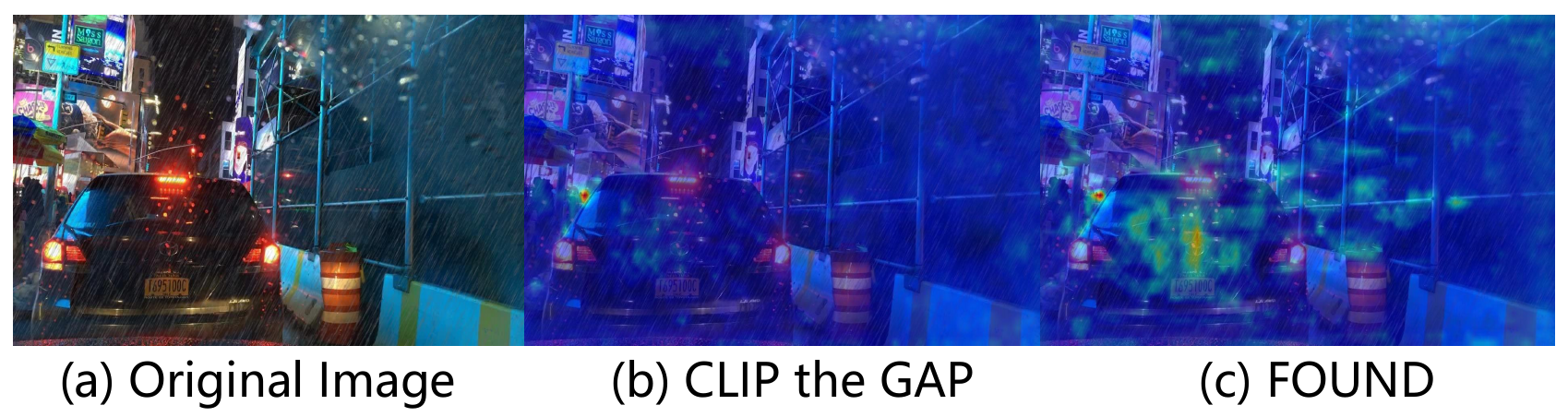}
    \caption{The baseline's feature map (b) is diffuse and distracted by background noise. In contrast, our FOUND model (c) generates a sharp activation map tightly focused on the target object.}
    \label{fig:feature_map}
\end{figure}

\subsection{Visualization and Analysis}
To provide deeper insights into our method's effectiveness, we visualize the learned feature representations in Figure~\ref{fig:tsne}, comparing the t-SNE projections of RoI features from the CLIP the Gap baseline and our FOUND framework. The baseline's feature space (left) exhibits less defined clusters and considerable overlap between object classes, indicating challenges in learning a fully discriminative representation. In stark contrast, FOUND (right) learns a remarkably well-structured space, forming distinct and compact clusters for each category. This demonstrates superior intra-class compactness and inter-class separability, a direct result of our von Mises-Fisher regularization. 


Beyond the global feature space, Figure~\ref{fig:feature_map} illustrates the model's feature-level attention in a challenging night-rainy scene. The baseline's feature map (b) is diffuse, with activations scattered across both the target car and irrelevant background noise like street signs and rain artifacts. Conversely, our model (c) generates a sharp, localized activation map tightly focused on the target object. This enhanced focus is achieved through our dual-component approach: PFA trains the model to be invariant to stylistic noise by perturbing the amplitude spectrum, while vMF regularization ensures that the learned representations remain robust. 




\section{Conclusion}
In this paper, we propose a novel method FOUND which is a robust framework for single-domain generalization in object detection. We identify and address the issue of semantic feature collapse caused by strong augmentation. Our solution integrates Probabilistic Fourier Augmentation to simulate domain shifts in the frequency domain with von Mises-Fisher regularization to maintain semantic feature structure. Experiments on challenging adverse-weather benchmarks show that FOUND consistently outperforms state-of-the-art methods, demonstrating superior generalization, especially in extreme conditions. 


{
    \small
    \bibliographystyle{ieeenat_fullname}
    \bibliography{main}

@article{Oppenheim1980TheIO,
  title={The importance of phase in signals},
  author={Alan V. Oppenheim and Jae Sung Lim},
  journal={Proceedings of the IEEE},
  year={1980},
  volume={69},
  pages={529-541},
  url={https://api.semanticscholar.org/CorpusID:23790114}
}

@inproceedings{choi2021robustnet,
  title={Robustnet: Improving domain generalization in urban-scene segmentation via instance selective whitening},
  author={Choi, Sungha and Jung, Sanghun and Yun, Huiwon and Kim, Joanne T and Kim, Seungryong and Choo, Jaegul},
  booktitle={Proceedings of the IEEE/CVF conference on computer vision and pattern recognition},
  pages={11580--11590},
  year={2021}
}

@article{volpi2018generalizing,
  title={Generalizing to unseen domains via adversarial data augmentation},
  author={Volpi, Riccardo and Namkoong, Hongseok and Sener, Ozan and Duchi, John C and Murino, Vittorio and Savarese, Silvio},
  journal={Advances in neural information processing systems},
  volume={31},
  year={2018}
}

@InProceedings{pmlr-v139-radford21a,
  title = 	 {Learning Transferable Visual Models From Natural Language Supervision},
  author =       {Radford, Alec and Kim, Jong Wook and Hallacy, Chris and Ramesh, Aditya and Goh, Gabriel and Agarwal, Sandhini and Sastry, Girish and Askell, Amanda and Mishkin, Pamela and Clark, Jack and Krueger, Gretchen and Sutskever, Ilya},
  booktitle = 	 {Proceedings of the 38th International Conference on Machine Learning},
  pages = 	 {8748--8763},
  year = 	 {2021},
  editor = 	 {Meila, Marina and Zhang, Tong},
  volume = 	 {139},
}

@inproceedings{qiao2020learning,
  title={Learning to learn single domain generalization},
  author={Qiao, Fengchun and Zhao, Long and Peng, Xi},
  booktitle={Proceedings of the IEEE/CVF conference on computer vision and pattern recognition},
  pages={12556--12565},
  year={2020}
}

@inproceedings{yun2019cutmix,
  title={Cutmix: Regularization strategy to train strong classifiers with localizable features},
  author={Yun, Sangdoo and Han, Dongyoon and Oh, Seong Joon and Chun, Sanghyuk and Choe, Junsuk and Yoo, Youngjoon},
  booktitle={Proceedings of the IEEE/CVF international conference on computer vision},
  pages={6023--6032},
  year={2019}
}

@article{hendrycks2019augmix,
  title={Augmix: A simple data processing method to improve robustness and uncertainty},
  author={Hendrycks, Dan and Mu, Norman and Cubuk, Ekin D and Zoph, Barret and Gilmer, Justin and Lakshminarayanan, Balaji},
  journal={arXiv preprint arXiv:1912.02781},
  year={2019}
}

@inproceedings{pan2018two,
  title={Two at once: Enhancing learning and generalization capacities via ibn-net},
  author={Pan, Xingang and Luo, Ping and Shi, Jianping and Tang, Xiaoou},
  booktitle={Proceedings of the european conference on computer vision (ECCV)},
  pages={464--479},
  year={2018}
}

@article{Huang2017ArbitraryST,
  title={Arbitrary Style Transfer in Real-Time with Adaptive Instance Normalization},
  author={Xun Huang and Serge J. Belongie},
  journal={2017 IEEE International Conference on Computer Vision (ICCV)},
  year={2017},
  pages={1510-1519},
  url={https://api.semanticscholar.org/CorpusID:6576859}
}

@article{Yang2020FDAFD,
  title={FDA: Fourier Domain Adaptation for Semantic Segmentation},
  author={Yanchao Yang and Stefano Soatto},
  journal={2020 IEEE/CVF Conference on Computer Vision and Pattern Recognition (CVPR)},
  year={2020},
  pages={4084-4094},
  url={https://api.semanticscholar.org/CorpusID:215745272}
}

@article{Khosla2020SupervisedCL,
  title={Supervised Contrastive Learning},
  author={Prannay Khosla and Piotr Teterwak and Chen Wang and Aaron Sarna and Yonglong Tian and Phillip Isola and Aaron Maschinot and Ce Liu and Dilip Krishnan},
  journal={ArXiv},
  year={2020},
  volume={abs/2004.11362},
  url={https://api.semanticscholar.org/CorpusID:216080787}
}

@inproceedings{zhang2024fine,
  title={Fine-grained Image-to-LiDAR Contrastive Distillation with Visual Foundation Models},
  author={Zhang, Yifan and Hou, Junhui},
  booktitle={Advances in Neural Information Processing Systems},
  year={2024}
}

@InProceedings{Xu_2021_CVPR,
author={Xu, Qinwei and Zhang, Ruipeng and Zhang, Ya and Wang, Yanfeng and Tian, Qi},
title={A Fourier-Based Framework for Domain Generalization},
booktitle={Proceedings of the IEEE/CVF Conference on Computer Vision and Pattern Recognition (CVPR)},
month={June},
year={2021},
pages={14383-14392}
}

@InProceedings{Vidit_2023_CVPR,
author={Vidit, Vidit and Engilberge, Martin and Salzmann, Mathieu},
title={CLIP the Gap: A Single Domain Generalization Approach for Object Detection},
booktitle={Proceedings of the IEEE/CVF Conference on Computer Vision and Pattern Recognition (CVPR)},
month={June},
year={2023},
pages={3219-3229}
}

@inproceedings{wu2022single,
title={Single-Domain Generalized Object Detection in Urban Scene via Cyclic-Disentangled Self-Distillation},
author={Wu, Aming and Deng, Cheng},
booktitle={Proceedings of the IEEE/CVF Conference on Computer Vision and Pattern Recognition},
pages={847--856},
year={2022}
}

@article{wang2024yolov10,
  title={Yolov10: Real-time end-to-end object detection},
  author={Wang, Ao and Chen, Hui and Liu, Lihao and Chen, Kai and Lin, Zijia and Han, Jungong and others},
  journal={Advances in Neural Information Processing Systems},
  volume={37},
  pages={107984--108011},
  year={2024}
}

@inproceedings{zhang2025style,
  title={Style Evolving along Chain-of-Thought for Unknown-Domain Object Detection},
  author={Zhang, Zihao and Wu, Aming and Han, Yahong},
  booktitle={Proceedings of the Computer Vision and Pattern Recognition Conference},
  pages={14225--14234},
  year={2025}
}

@inproceedings{wu2025percept,
  title={Percept, Memory, and Imagine: World Feature Simulating for Open-Domain Unknown Object Detection},
  author={Wu, Aming and Deng, Cheng},
  booktitle={Proceedings of the Computer Vision and Pattern Recognition Conference},
  pages={4682--4691},
  year={2025}
}

@inproceedings{you2019universal,
  title={Universal domain adaptation},
  author={You, Kaichao and Long, Mingsheng and Cao, Zhangjie and Wang, Jianmin and Jordan, Michael I},
  booktitle={Proceedings of the IEEE/CVF Conference on Computer Vision and Pattern recognition},
  pages={2720--2729},
  year={2019}
}

@article{ben2006analysis,
  title={Analysis of representations for domain adaptation},
  author={Ben-David, Shai and Blitzer, John and Crammer, Koby and Pereira, Fernando},
  journal={Advances in neural information processing systems},
  volume={19},
  year={2006}
}

@inproceedings{ganin2015unsupervised,
  title={Unsupervised domain adaptation by backpropagation},
  author={Ganin, Yaroslav and Lempitsky, Victor},
  booktitle={International conference on machine learning},
  pages={1180--1189},
  year={2015},
}

@inproceedings{wang2021learning,
  title={Learning to diversify for single domain generalization},
  author={Wang, Zijian and Luo, Yadan and Qiu, Ruihong and Huang, Zi and Baktashmotlagh, Mahsa},
  booktitle={Proceedings of the IEEE/CVF International Conference on Computer Vision},
  pages={834--843},
  year={2021}
}

@inproceedings{li2025seeking,
  title={Seeking consistent flat minima for better domain generalization via refining loss landscapes},
  author={Li, Aodi and Zhuang, Liansheng and Long, Xiao and Yao, Minghong and Wang, Shafei},
  booktitle={Proceedings of the Computer Vision and Pattern Recognition Conference},
  pages={15349--15359},
  year={2025}
}

@inproceedings{agarwal2025tide,
  title={TIDE: Training Locally Interpretable Domain Generalization Models Enables Test-time Correction},
  author={Agarwal, Aishwarya and Karanam, Srikrishna and Gandhi, Vineet},
  booktitle={Proceedings of the Computer Vision and Pattern Recognition Conference},
  pages={30210--30220},
  year={2025}
}

@article{li2025phrase,
  title={Phrase grounding-based style transfer for single-domain generalized object detection},
  author={Li, Hao and Wang, Wei and Wang, Cong and Wang, Mengzhu and Zhang, Xiang and Lan, Long and Liu, Xinwang and Li, Kenli and Cao, Xiaochun},
  journal={IEEE Transactions on Circuits and Systems for Video Technology},
  year={2025},
}

@inproceedings{yuan2025asgs,
  title={ASGS: Single-Domain Generalizable Open-Set Object Detection via Adaptive Subgraph Searching},
  author={Yuan, Yuxuan and Tang, Luyao and Chen, Yixin and Chen, Chaoqi and Huang, Yue and Ding, Xinghao},
  booktitle={Proceedings of the IEEE/CVF International Conference on Computer Vision},
  pages={20911--20921},
  year={2025}
}

@article{ren2015faster,
  title={Faster r-cnn: Towards real-time object detection with region proposal networks},
  author={Ren, Shaoqing and He, Kaiming and Girshick, Ross and Sun, Jian},
  journal={Advances in neural information processing systems},
  volume={28},
  year={2015}
}

@inproceedings{pan2019switchable,
  title={Switchable whitening for deep representation learning},
  author={Pan, Xingang and Zhan, Xiaohang and Shi, Jianping and Tang, Xiaoou and Luo, Ping},
  booktitle={Proceedings of the IEEE/CVF international conference on computer vision},
  pages={1863--1871},
  year={2019}
}

@inproceedings{huang2019iterative,
  title={Iterative normalization: Beyond standardization towards efficient whitening},
  author={Huang, Lei and Zhou, Yi and Zhu, Fan and Liu, Li and Shao, Ling},
  booktitle={Proceedings of the IEEE/CVF conference on computer vision and pattern recognition},
  pages={4874--4883},
  year={2019}
}

@inproceedings{wang2025dynamic,
  title={Dynamic-vlm: Simple dynamic visual token compression for videollm},
  author={Wang, Han and Nie, Yuxiang and Ye, Yongjie and Wang, Yanjie and Li, Shuai and Yu, Haiyang and Lu, Jinghui and Huang, Can},
  booktitle={Proceedings of the IEEE/CVF International Conference on Computer Vision},
  pages={20812--20823},
  year={2025}
}

@inproceedings{kim2024vlm,
  title={Vlm-pl: Advanced pseudo labeling approach for class incremental object detection via vision-language model},
  author={Kim, Junsu and Ku, Yunhoe and Kim, Jihyeon and Cha, Junuk and Baek, Seungryul},
  booktitle={Proceedings of the IEEE/CVF Conference on Computer Vision and Pattern Recognition},
  pages={4170--4181},
  year={2024}
}

@article{wang2024q,
  title={Q-vlm: Post-training quantization for large vision-language models},
  author={Wang, Changyuan and Wang, Ziwei and Xu, Xiuwei and Tang, Yansong and Zhou, Jie and Lu, Jiwen},
  journal={Advances in Neural Information Processing Systems},
  volume={37},
  pages={114553--114573},
  year={2024}
}

@inproceedings{danish2024improving,
  title={Improving single domain-generalized object detection: A focus on diversification and alignment},
  author={Danish, Muhammad Sohail and Khan, Muhammad Haris and Munir, Muhammad Akhtar and Sarfraz, M Saquib and Ali, Mohsen},
  booktitle={Proceedings of the IEEE/CVF Conference on Computer Vision and Pattern Recognition},
  pages={17732--17742},
  year={2024}
}

@article{parchami2016recent,
  title={Recent developments in speech enhancement in the short-time Fourier transform domain},
  author={Parchami, Mahdi and Zhu, Wei-Ping and Champagne, Benoit and Plourde, Eric},
  journal={IEEE Circuits and Systems Magaz},
  volume={16},
  number={3},
  pages={45--77},
  year={2016},
}

@article{tao2008sampling,
  title={Sampling and sampling rate conversion of band limited signals in the fractional Fourier transform domain},
  author={Tao, Ran and Deng, Bing and Zhang, Wei-Qiang and Wang, Yue},
  journal={IEEE Transactions on Signal Processing},
  volume={56},
  number={1},
  pages={158--171},
  year={2008},
}

@article{banerjee2005clustering,
  title={Clustering on the Unit Hypersphere using von Mises-Fisher Distributions.},
  author={Banerjee, Arindam and Dhillon, Inderjit S and Ghosh, Joydeep and Sra, Suvrit and Ridgeway, Greg},
  journal={Journal of Machine Learning Research},
  volume={6},
  number={9},
  year={2005}
}

@article{du2024probabilistic,
  title={Probabilistic contrastive learning for long-tailed visual recognition},
  author={Du, Chaoqun and Wang, Yulin and Song, Shiji and Huang, Gao},
  journal={IEEE Transactions on Pattern Analysis and Machine Intelligence},
  volume={46},
  number={9},
  pages={5890--5904},
  year={2024},
}

@article{du2024simpro,
  title={Simpro: A simple probabilistic framework towards realistic long-tailed semi-supervised learning},
  author={Du, Chaoqun and Han, Yizeng and Huang, Gao},
  journal={Neural Information Processing Systems},
  year={2024}
}

@article{van2014renyi,
  title={R{\'e}nyi divergence and Kullback-Leibler divergence},
  author={Van Erven, Tim and Harremos, Peter},
  journal={IEEE Transactions on Information Theory},
  volume={60},
  number={7},
  pages={3797--3820},
  year={2014},
}

@article{cui2024decoupled,
  title={Decoupled kullback-leibler divergence loss},
  author={Cui, Jiequan and Tian, Zhuotao and Zhong, Zhisheng and Qi, Xiaojuan and Yu, Bei and Zhang, Hanwang},
  journal={Advances in Neural Information Processing Systems},
  volume={37},
  pages={74461--74486},
  year={2024}
}

@inproceedings{yu2020bdd100k,
  title={Bdd100k: A diverse driving dataset for heterogeneous multitask learning},
  author={Yu, Fisher and Chen, Haofeng and Wang, Xin and Xian, Wenqi and Chen, Yingying and Liu, Fangchen and Madhavan, Vashisht and Darrell, Trevor},
  booktitle={Proceedings of the IEEE/CVF Conference on Computer Vision and Pattern Recognition},
  pages={2636--2645},
  year={2020}
}

@inproceedings{cordts2016cityscapes,
  title={The cityscapes dataset for semantic urban scene understanding},
  author={Cordts, Marius and Omran, Mohamed and Ramos, Sebastian and Rehfeld, Timo and Enzweiler, Markus and Benenson, Rodrigo and Franke, Uwe and Roth, Stefan and Schiele, Bernt},
  booktitle={Proceedings of the IEEE Conference on Computer Vision and Pattern Recognition},
  pages={3213--3223},
  year={2016}
}

@article{rao2024srcd,
  title={Srcd: Semantic reasoning with compound domains for single-domain generalized object detection},
  author={Rao, Zhijie and Guo, Jingcai and Tang, Luyao and Huang, Yue and Ding, Xinghao and Guo, Song},
  journal={IEEE Transactions on Neural Networks and Learning Systems},
  year={2024},
}

@article{hassaballah2020vehicle,
  title={Vehicle detection and tracking in adverse weather using a deep learning framework},
  author={Hassaballah, Mahmoud and Kenk, Mourad A and Muhammad, Khan and Minaee, Shervin},
  journal={IEEE Transactions on Intelligent Transportation Systems},
  volume={22},
  number={7},
  pages={4230--4242},
  year={2020},
}

@article{sakaridis2018semantic,
  title={Semantic foggy scene understanding with synthetic data},
  author={Sakaridis, Christos and Dai, Dengxin and Van Gool, Luc},
  journal={International Journal of Computer Vision},
  volume={126},
  number={9},
  pages={973--992},
  year={2018},
}
}

\end{document}